\documentclass[letterpaper, 10 pt, conference]{ieeeconf}  

\IEEEoverridecommandlockouts                              

\usepackage{graphics} 
\usepackage{epsfig} 
\usepackage{mathptmx} 
\usepackage{times} 
\usepackage{amsmath} 
\usepackage{amssymb}  
\usepackage{multirow}
\usepackage{hyperref}

\usepackage{booktabs} 

\usepackage{ntheorem} 
\theoremseparator{:} 

\newtheorem*{hyp*}{Hypothesis \protect\hypnumber} 

\newcommand{\hypnumber}{}

\usepackage{todonotes} 

\title{\LARGE \bf

Towards Driving Policies with Personality: Modeling Behavior and Style in Risky Scenarios via Data Collection in Virtual Reality
}

\author{Laura Zheng, Julio Poveda, James Mullen, Shreelekha Revankar, Ming C. Lin
\thanks{The authors are affiliated with the Department of Computer Science,
        University of Maryland, College Park} \\
        \href{https://gamma.umd.edu/vrdrivingrisk}{\texttt{gamma.umd.edu/vrdrivingrisk}}
}

\begin{document}

\maketitle
\thispagestyle{empty}
\pagestyle{empty}

\begin{abstract}
Autonomous driving research currently faces data sparsity in representation of risky scenarios. Such data is both difficult to obtain ethically in the real world, and unreliable to obtain via simulation. Recent advances in virtual reality (VR) driving simulators lower barriers to tackling this problem in simulation. We propose the first data collection framework for risky scenario driving data from real humans using VR, as well as accompanying numerical driving personality characterizations. We validate the resulting dataset with statistical analyses and model driving behavior with an eight-factor personality vector based on the Multi-dimensional Driving Style Inventory (MDSI). Our method, dataset, and analyses show that realistic driving personalities can be modeled without deep learning or large datasets to complement autonomous driving research.

\end{abstract}

\section{Introduction}

Ideally, autonomous vehicles should be much safer than human-driven vehicles. However, risk identification, decision making, and avoidance are all difficult and data-sparse challenges in autonomous driving research. 
To avoid vehicle collisions, autonomous driving frameworks require accurate trajectory forecasting and driver profiling. Interactions between actors (e.g., pedestrians and other drivers) can vary greatly and may not be easily predictable. While complex behavior can be learned implicitly end-to-end with deep learning, we argue that personality can be modeled directly with respect to sensor data; this approach is not only driven by data from real people, but also avoids generalization and distributional shift drawbacks associated with deep learning for autonomous driving. 

We can address risk avoidance in driving as long as we have a sufficiently annotated and well-sized dataset. However, there are several challenges to this. First, risky and dangerous scenarios are expensive and unethical to collect in the real world, especially when equipped with sensor annotations. Secondly, collecting risky and dangerous scenarios in simulation may not produce accurate or reliable data, especially with sim-to-real domain transfer as an additional step to practical use in the real world. The co-existence of both challenges creates a deadlock in advancement of risk handling for autonomous driving research. 

With recent developments in virtual reality (VR) systems for autonomous driving testbed simulators, conducting user studies for autonomous driving research is now more accessible and easily applicable to existing autonomous driving frameworks. In this paper, we tackle the lack of human data in simulation to break the deadlock on risky driving data collection. With annotated data and analysis available, autonomous driving researchers can use data-driven assumptions to account for complex human behavior.

\begin{figure}[t!]
    \centering
    \includegraphics[width=8cm]{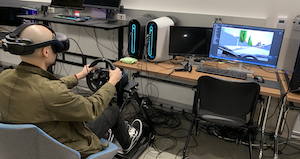}
    \caption{\textbf{VR data collection setup.} A pilot user drives through a test scenario. }
    \label{fig:teaser}
    \vspace{-1em}
\end{figure}

In this paper, we design and implement the first VR data collection framework for driving behavior in risky scenarios and model driving behavior with intuitive and interpretable driving personality characterization. We then validate our collected data using statistical methods and provide insight on relationships between personality factors and measured driving behavior. Moreover, we show how to relate rich driving sensor data to eight interpretable personality factors.

Personality factors are determined by the Multi-dimensional Driver Style Inventory (MDSI)~\cite{Taubman-Ben-Ari_Mikulincer_Gillath_2004}. The MDSI is a well-validated and established driver style personality questionnaire which characterizes a user's driving style into eight distinct factors: reckless, anxious, risky, angry, high-velocity, distress-reducing, patient, and careful. 
Correlation coefficients and p-values from statistical tests can be used to model relationships between a desired driving personality and control input and trajectory metrics. For example, our correlation analysis shows that ``patient" driving style can be modelled confidently as a \textit{linear} function of steering, brake, and perhaps even throttle, in order of priority. This relationship can then be applied as a linear constraint in learning for a driving policy. 
This work also serves as a practical guide in conducting 
VR studies for safer autonomous driving. 
The key contributions are: 
\begin{enumerate}
    \item A novel VR data collection framework for human driving in simulators; 
    \item An autonomous driving dataset with personality annotations and sensor measurements; 
    \item Correlation analyses which can be directly applied to learning policies as constraints for simulating driving personality; 
    \item A model representing the relationship between driving data and personality factors, enabling characterization of non-human driving agents.
\end{enumerate}

\begin{figure*}
    \centering
  \includegraphics[width=15cm, height=6cm]{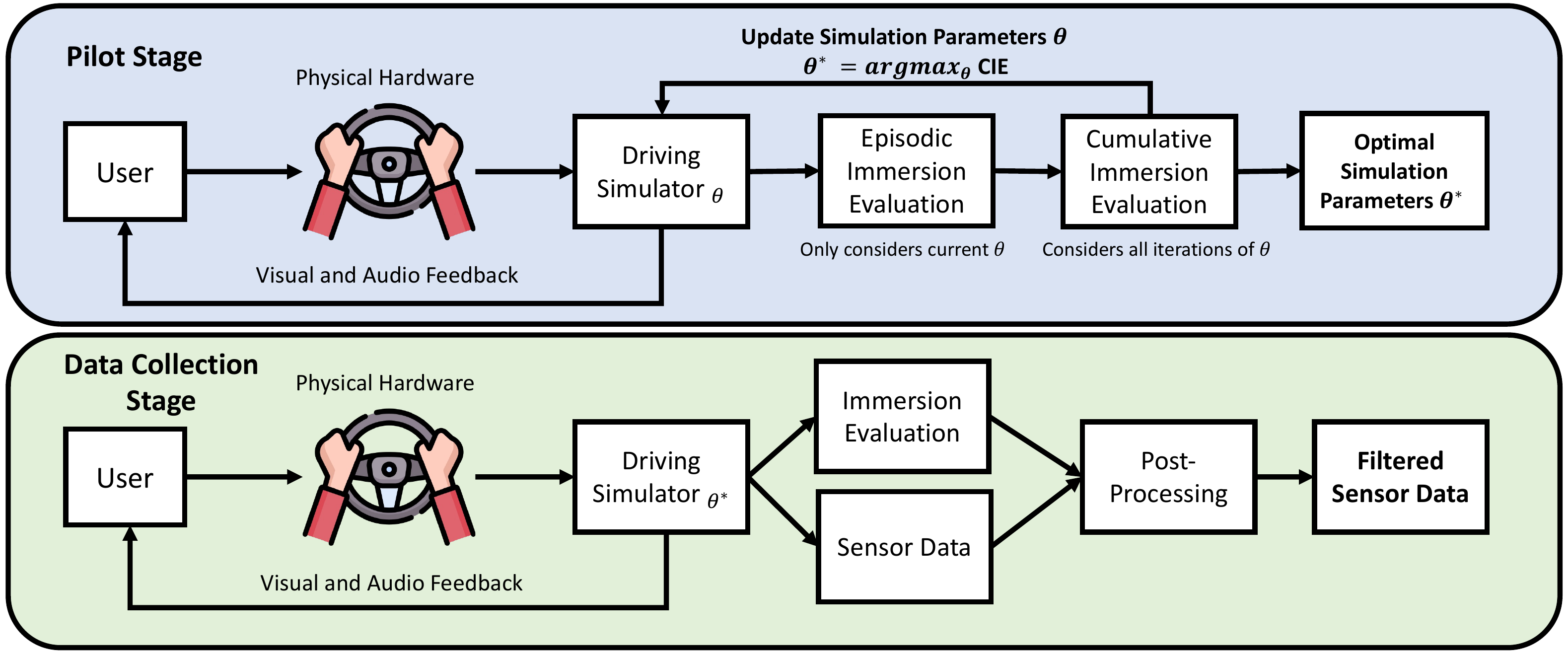}
  \vspace*{-0.5em}
  \caption{\textbf{System diagram for VR human subject data collection.} Pictured above is the iterative improvement scheme of our VR user study in driving in two stages. During the pilot stage (in blue), episodic and cumulative immersion evaluations are used to assess the user experience of the VR Driving Simulator, which is parameterized by $\theta$. Driving simulator parameters $\theta$ include visual elements, audio, frame rate, rendering effects such as mirrors, etc. Due to constraints of real-time simulation, in addition to hardware constraints, it is nontrivial to tune simulation parameters for maximum immersion. Optimal driving simulation parameters $\theta^{*}$ are then used in a larger-scale data collection stage. }
  \label{fig:control_system}
\end{figure*}

\section{Related Works}

\subsection{Users studies on VR driving}
For decades, user studies have been an essential method for humans to identify and understand people's needs in various contexts \cite{wilson1981}. These types of studies have been conducted by researchers from many knowledge domains, including privacy and security \cite{zeng2019}, 
mHealth \cite{kim2017}, 
among others. 
In terms of VR driving, there are a few existing works in user studies.
For instance, Weidner et al. conducted a user study to compare people's physiological data and driving performance when using two displays for driving simulators, one using a VR head-mounted display and another one without it~\cite{weidner2017}. In another user study, Yeo et al. compared six self-driving simulators to identify which had higher fidelity and sense of presence~\cite{yeo2020}. Similarly, Lang et al. did a study to compare driving training methods, two of which involved the use of VR~\cite{lang2018}.

\subsection{Modeling Driving Behavior}
While much research in driving and transportation tackle system-related problems, another sector deals with modeling and addressing the complexity of human behavior. Formulaic traffic representations generally oversimplify human behavior by modeling systems with a single ordinary differential equation~\cite{Treiber_Hennecke_Helbing_2000, Newell_2002, Wiedemann_1974, krauss}. In this section, we discuss various prior work in driving and transportation that seeks to model or analyze complexity of human driving behavior, characterizing factors such as comfort in driving, familiarity with location, weather and road conditions, emotional state, traffic conditions, etc.
Some works have explored driving personality from real-world data collection; one such example is the METEOR driving dataset. Chandra et al. consolidated this dataset with annotations including rare and interesting driving behaviors such as cut-ins, yielding, overtaking, overspeeding, zigzagging, and rule-breaking, etc~\cite{chandra2021meteor}. B-GAP by Mavrogiannis et al. explores the procedural generation of diverse driving behaviors~\cite{mavrogiannis2020bgap}. An earlier work by Schwarting et al. describe driving behavior with a Social Value Orientation (SVO) metric, which quantifies a driver’s degree of selfishness or altruism~\cite{Schwarting_Pierson_Alonso-Mora_Karaman_Rus_2019}. SVO is then validated through experiments involving simulation as well as real world traffic trajectory data. Moreover, Yan et al. conducted a simulated driving experiment to study the correlation between driving behavior, personality, and electroencephalography~\cite{yan2019}. 
Participants' personalities were identified by applying the Cattell 16 Personality Factor Questionnaire~\cite{cattell2008} and via hierarchical clustering of the 16 factors participants were categorized into one of three groups: Insensitivity, Apprehension, or Unreasoning. Their results demonstrate a strong correlation between the three main variables studied (driving behavior, personality, and electroencephalography).

In a review of Virtual Reality Applications for Automated Driving from 2009 to 2020, Riegler et al. found that several existing behavioral studies do not 1) emphasize enough the importance of pilot studies, 2) establish heuristics or standardizations for behavioral studies in the automotive domain, 3) involve a highly realistic, context-rich real world representation in VR, otherwise mentioned as a “digital twin”~\cite{Riegler_Riener_Holzmann_2021}. Some common quantitative metrics collected by behavioral studies in driving include distance, steering, acceleration/braking, and gaze metrics, while subjective data included workload (NASA TLX), self-ratings, and SSQ (simulator sickness questionnaire). 
Most recently, a driving simulator built for behavioral research, DReyeVR~\cite{Silvera_Biswas_Admoni_2022}, was 
presented -- building upon the well-established CARLA driving simulator with added data collection, VR functionality, and hardware integration. 
DReyeVR's compatibility with CARLA makes VR user studies for driving democratized and thus opens doors for research which is compatible with many existing state-of-the-art autonomous driving methods, especially those built off Learning By Cheating~\cite{chen2019lbc, Prakash2021CVPR, Chitta2022PAMI}. 

\begin{figure*}[ht!]
    \centering
    \includegraphics[width=\textwidth]{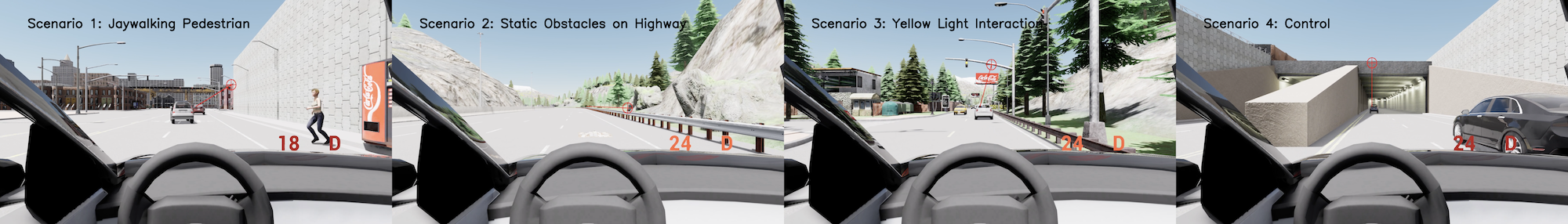}
    \vspace{-1.5em}
    \caption{Previews of Scenario 1-4, as described in TABLE~\ref{tab:scenario_descriptions}, in order from left to right. }
    \label{fig:scenarios_screenshots}
    \vspace{-0.5em}
\end{figure*}

\begin{table}[ht!]
\caption{VR driving scenario descriptions. \\}
\centering
\begin{tabular}{c p{0.8\linewidth}}
\toprule
Scenario & Description \\
\midrule 
1 & 2-lane straight urban city road, about 60 meters away from a traffic intersection. Nearby traffic is moderately populated. When the user drives about 40 meters forward, a \textit{jaywalking pedestrian} begins to run across the street from behind a vending machine. The scenario ends once the user reaches the traffic intersection. \\
2 & 4-lane highway, with no nearby traffic at all. When the user drives about 100 meters forward, three static piles of garbage appear one after another in the lane. If the vehicle drives over the obstacle, the user may feel as though they are going over a bump. Scenario ends after about 200 meters. \\
3 & Straight suburban road, surrounded by trees, about 50 meters away from a traffic intersection. Once the user drives near the traffic intersection, the traffic light turns from green to yellow. The scenario ends after the user crosses the intersection. \\
4 & 2-lane urban city road, slightly curved. No particular risks are presented in this scenario. Traffic density is slight, with two other vehicles around the user. The two vehicles are in different lanes and each going at slow speed, giving the user the options to stay behind a leading vehicle or change lanes and pass. The scenario ends after the user drives through an underpass and stops at the next traffic intersection. \\
\bottomrule
\end{tabular}
\label{tab:scenario_descriptions}
\vspace{-1em}
\end{table}

\section{Methodology}

\subsection{Human Subject Data Collection}

\textbf{Format.}
The user's data collection study has 3 components: a preliminary questionnaire, the VR component, then a post-VR questionnaire. The project website includes details of these components and why they are warranted.

\textbf{Participant Briefing.}
Upon arrival at the study location, participants are greeted and briefed on expectations on the study and their rights. 
Before beginning the study, participants will need to read and sign an informed consent form. 

\textbf{Preliminary Questionnaire.}
Users are asked to fill out a preliminary questionnaire that served a few purposes. Firstly, the questionnaire collects optional demographic information provided by the user. Second, it measures the user's self-reported physiological state before exposure to VR for the purpose of measuring VR sickness later; these questions are standard questions from the Simulator Sickness Questionnaire (SSQ)~\cite{ssq} commonly used in VR user studies. Lastly, the questionnaire includes a personality test to determine driving traits of the user. The personality test is the Multidimensional Driving Style Inventory (MDSI)~\cite{Taubman-Ben-Ari_Mikulincer_Gillath_2004}, a validated 44-item questionnaire commonly used as a standard in transportation research.  
Through MDSI analysis, driving personality can be factorized into eight characteristics: dissociative, anxious, risky, angry, high-velocity, distress reduction, patient, and careful.
Upon completion of this component, users are characterized by a weighted sum of these characteristics, which are later used for analysis with the collected data from the driving tasks our users completed.

\begin{table}
\small
\caption[Demographics]{
        Participant demographics 
        (n=24)
    }
\vspace{-4mm}
\centering
\begin{tabular}{llr}
\toprule
Gender & Male &  66.7\% \\
          & Female &  29.2\% \\
          & Non-binary &   4.2\% \\
Ethnicity & Asian &  62.5\% \\
          & Black/African &  8.3\% \\
          & Caucasian & 25\% \\
          & Armenian/Jewish & 4.2\% \\
          & South East Asian & 4.2\% \\
Age group & 18-29 &  83.3\% \\
          & 30-45 &  16.7\% \\
\bottomrule
\end{tabular}
\label{tab:demographics}

\label{tab:demographics}
\vspace{-8mm}
\end{table}

\textbf{VR Scenarios.}
The main component of the study is measuring user decision making in VR in response to pre-crash scenarios. Urban driving involves many unpredictable and moving factors, such as high traffic density, erratic pedestrian behavior, and frequent violation of traffic laws. 
The NHTSA established a report on frequent pre-crash scenarios in 2013~\cite{NHTSA}. These scenarios detail the conditions leading up to light-vehicle crashes, with specific examples including running red lights, rear ending, loss of control, and lane changing. Recent developments in driving simulators make it possible to simulate such pre-crash scenarios and port them to VR. In this section, we choose four urban scenarios based on common autonomous driving benchmarks and the NHTSA pre-crash typology to present to the user. An additional fifth scenario, which is longer and randomized, is presented to the user for dataset collection purposes, however is not used for behavior analysis in the results.
Collecting data from more than four short scenarios from human subjects is difficult due to constraints related to participant attention span and fatigue. 


The four short scenarios include a control scenario with no pre-crash conditions, jaywalking pedestrian, yellow light interaction, and a static road obstacle. 
Scenarios are non-traumatic, as collisions and ragdoll physics are unrealistic and low-speed. This does not affect the realism of user data, so long as the user is immersed and expects realistic interactions prior to the interaction. In addition, the user experiences no acceleration or collision forces in the real world.
Each scenario is presented to the user in a random order to mitigate any potential biases in the data that may arise from ordering effects.
Since our study has four scenarios, we aim to have at least 24 participants in our data. 
Examples of sensor data collected are trajectory, VR camera rotation, and control inputs like steering and throttle. Other modalities of data commonly available in CARLA such as RGB, lidar, and depth images can be obtained post-hoc with simulation recording logs. 
Each scenario is first driven by an autopilot agent to demonstrate the intended driving route with all risks removed. This is also to build the expectation of a smooth driving route for the user prior to introducing unexpected factors. 
Then, the user will drive the route themselves, except this time with risks or unexpected behavior added in. 
Scenarios are also intentionally simple in route, involving no turns and representing mostly straight road driving. Detailed descriptions of each scenario are presented in Table~\ref{tab:scenario_descriptions}. 
During each run by the user, data such as trajectory, control input, eye-tracking sensor data, and RGB dashcam images are stored for future analysis. Qualitative observations are also noted by the interviewer. Upon completing the fourth scenario, the user now moves on to the post-VR questionnaire.

\begin{figure}[t!]
    \centering
    \includegraphics[width=0.9\columnwidth]{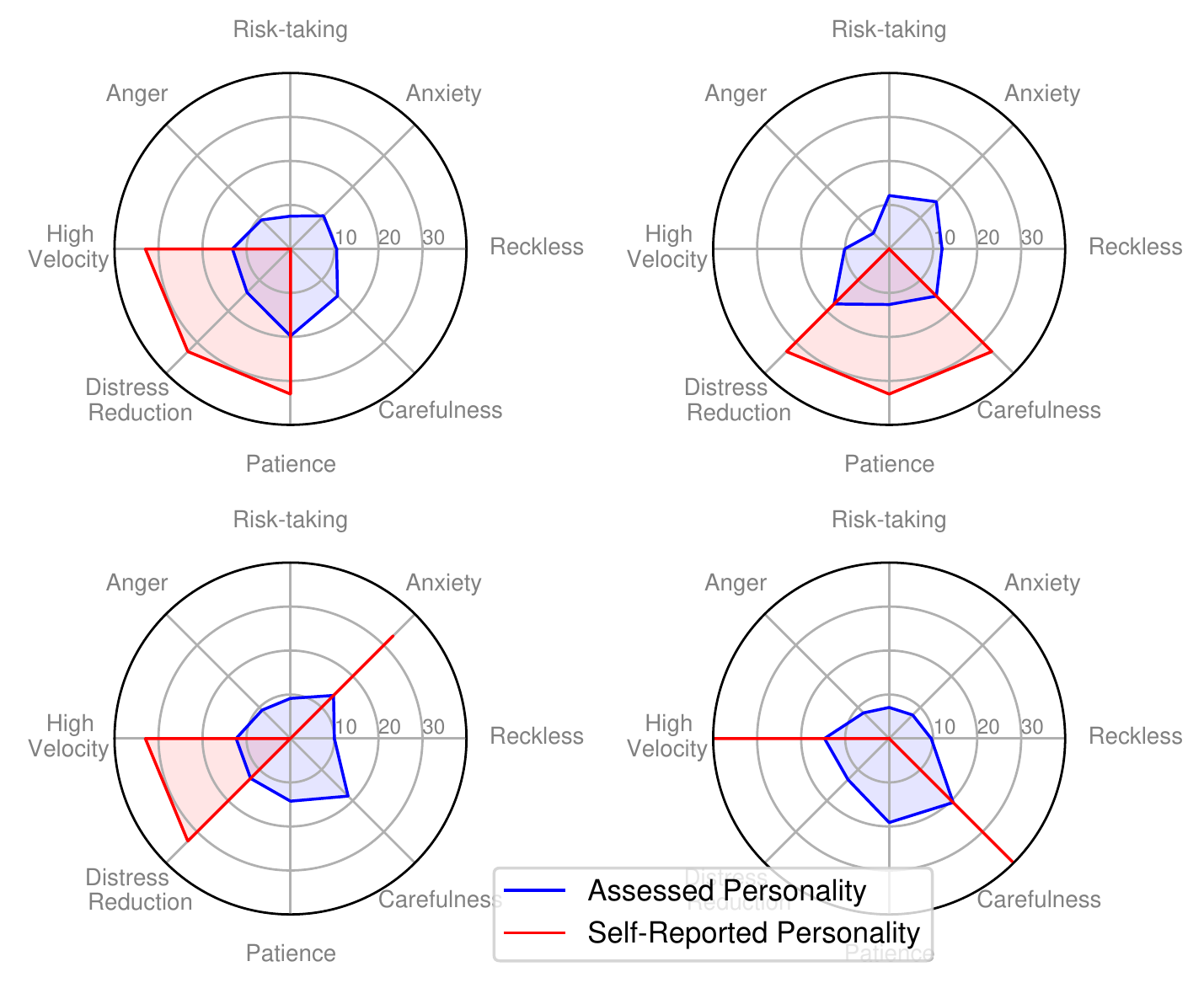}
    \vspace{-1em}
    \caption{MDSI personality results (blue) plotted against self-classifications of driving personality (red) of subjects, for four randomly selected samples. While self-classifications cannot be directly compared with MDSI results in value, we still observe differences in general monotonicity correlation. For example, the bottom left subject did not consider themselves a careful driver, but MDSI results indicate higher carefulness relative to other factors. }
    \label{fig:self_eval_vs_model}
    \vspace{-1.5em}
\end{figure}

\textbf{Post-VR Questionnaire.}
The user first evaluates their physiological state, similarly to the preliminary questionnaire. Then, they fill out questions related to the System Usability Scale (SUS) in order to measure the level of immersion. Poor immersion can be a detrimental factor to the fidelity of collected data in VR, as users may end up driving differently than they would in the real world if not fully immersed. 

\textbf{Recruitment and Outreach.}
We designed and shared a recruitment poster via various means in our university: (1) in an undergraduate newsletter, (2) in the CS graduate students Slack channel, and (3) in an email list to CS graduate students. Our poster had a link that redirected potential participants to a screening survey that asked them questions such as if they have strong susceptibility to motion sickness.

Next, a researcher manually checked the responses and selected eligible participants based on their responses (e.g., people aged 18 years or older who have a driving license). Table~\ref{tab:demographics} shows our participants' demographic information.

\textbf{Iterative Improvement and Pilot Study. } Since there are few precedents for this particular method of data collection, we conducted a pilot study to gain feedback on data collection study design. The pilot study improves user experience and simulation settings based on qualitative feedback, which is important for immersion and behavior realism. Since IRB approval was not obtained prior to the pilot studies, no results based on pilot studies will be presented. Details on this process is illustrated in Figure~\ref{fig:control_system}. 

\textbf{Ethical Considerations.}
Our data collection study was approved by our university's Institutional Review Board (IRB). Moreover, we took additional measures to protect participants and their data.
For instance, we reassured users that they could stop participating at any point, measured participants' inter-pupillary distance and adjusted the VR headset accordingly to make them have a comfortable VR experience.
Additionally, all participants' data was stored in a cloud repository that was only accessible to the researchers.

\textbf{Hardware.} We ran our experiments on a single computer with 11th Gen Intel Core processor, which has 8 cores, 32 GB of memory, and an NVIDIA RTX 3090 graphics card. 
The headset used is the Vive Focus 3, which is both portable and includes eye tracking sensors. The driving setup includes the Logitech G920 Driving Force Racing Wheel and Floor Pedals, which is attached to a driving setup stand. 

\textbf{Software.} For the VR driving scenarios, we use the DReyeVR simulator~\cite{Silvera_Biswas_Admoni_2022}, which ports the CARLA driving simulator~\cite{CARLA} to VR and provides useful tools for user studies. We built this simulator from source on Windows 10 and use the Scenario Runner library from CARLA to design experiments.

\begin{table}[t!]
\caption{ We validate our model from Section~\ref{sec:method_model} with K-means cross validation, in which a model is trained $K$ times on $K$ subsets and validated on different subsets of the data. Our model achieves a low validation mean squared error loss averaged over all folds. \\}
\centering
\begin{tabular}{ccc}
\toprule
Scenario & $\downarrow$ Avg Loss$_{Train}$ & $\downarrow$ Avg Loss$_{Valid}$ \\
\midrule 
1 & 0.0011 &  0.0023 \\
2 & 0.0008 &  0.0010 \\
3 & 0.0010 &  0.0014 \\
4 & 0.0008 &  0.0016 \\
\bottomrule
\end{tabular}
\label{tab:kmeans_validation}
\vspace{-1em}
\end{table}

\begin{table*}[ht!]
\caption{Pearson correlation table between personality components from MDSI~\cite{Taubman-Ben-Ari_Mikulincer_Gillath_2004} and sensor measurements for scenario 2. In this table, we find nine significant correlations (in bold), suggesting that some relationships between personality scores and measured driving behavior are non-spurious and can be modelled with rigorous linear assumptions. For example, researchers looking to model patient driving in driving policies may simulate patient driving behavior as a linear function of steering movement with high confidence. Other relationships with lower confidences, or higher p-values, can still be modelled in driving policy objectives; p-values can be used as weights in establishing objective priority.}
\centering
\begin{tabular}{lc|ccc|ccccc}
\toprule
& & \multicolumn{3}{|c|}{Control Input} & \multicolumn{5}{|c}{Trajectory Metrics: Scenario 2}\\
Score & & $\sum$ Steer &  $\sum$ Throttle & $\sum$ Brake & Sinuosity & Curvature$_{TA}$ & Curvature$_{LV}$ & Curvature$_{Steiner}$ & Curvature$_{OC}$ \\
\hline\hline
\multirow{ 3}{*}{Reckless} & $r_{Pearson}$ & 0.341 & -0.239 & -0.402 & -0.08 & 0.368 & 0.387 & 0.046 & -0.391 \\
& $p$ &  0.181 & 0.356 & 0.109 & 0.761 & 0.147 & 0.125 & 0.861 & 0.121 \\
\midrule 
\multirow{ 3}{*}{Anxious} & $r_{Pearson}$ & 0.01 & -0.213 & -0.275 & 0.212 & -0.438 & -0.454 & -0.085 & \textbf{-0.591} \\
          & $p$ & 0.971 & 0.412 & 0.285 & 0.415 & 0.079 & 0.067 & 0.745 & \textbf{0.012} \\
\midrule
\multirow{ 3}{*}{Risky} & $r_{Pearson}$ &  0.219 & -0.169 & -0.205 & -0.275 & 0.357 & 0.402 & -0.091 & \textbf{0.547} \\
          & $p$ &  0.397 & 0.516 & 0.430 & 0.285 & 0.16 & 0.109 & 0.729 & \textbf{0.023} \\
\midrule
\multirow{ 3}{*}{Angry} & $r_{Pearson}$ & -0.046 & 0.087 & 0.258 & 0.005 & \textbf{0.694} & \textbf{0.612} & \textbf{0.598} & 0.370 \\
          & $p$ &  0.860 & 0.74 & 0.318 & 0.986 & \textbf{0.002} & \textbf{0.009} & \textbf{0.011} & 0.144 \\
\midrule
\multirow{ 3}{*}{High-Velocity} & $r_{Pearson}$ & 0.147 & 0.001 & 0.242 & -0.246 & 0.278 & 0.206 & \textbf{0.511} & -0.026 \\
          & $p$ &  0.574 & 0.996 & 0.349 & 0.341 & 0.281 & 0.428 & \textbf{0.036} & 0.921 \\
\midrule
\multirow{ 3}{*}{Distress-Reducing} & $r_{Pearson}$ & -0.011 & -0.192 & -0.114 & -0.066 & -0.434 & -0.399 & -0.284 & 0.054 \\
          & $p$ &  0.967 & 0.459 & 0.663 & 0.8 & 0.082 & 0.112 & 0.269 & 0.836 \\
\midrule
\multirow{ 2}{*}{Patient} & $r_{Pearson}$ & \textbf{-0.539} & 0.461 & \textbf{0.492} & 0.288 & -0.102 & -0.133 & 0.108 & 0.094 \\
          & $p$ & \textbf{0.026} & 0.063 & \textbf{0.045} & 0.263 & 0.696 & 0.611 & 0.679 & 0.719 \\
\midrule
\multirow{ 3}{*}{Careful} & $r_{Pearson}$ & 0.079 & \textbf{0.501} & 0.193 & 0.113 & 0.043 & 0.095 & -0.339 & -0.127 \\
          & $p$ & 0.762 & \textbf{0.040} & 0.459 & 0.665 & 0.87 & 0.717 & 0.183 & 0.626 \\
\bottomrule
\end{tabular}

\label{tab:correlation_pearson}
\end{table*}

\begin{table*}[ht!]
\caption{Spearman correlation table between personality components from MDSI~\cite{Taubman-Ben-Ari_Mikulincer_Gillath_2004} and sensor measurements for scenario 2. This table demonstrates eight significant correlations (in bold) for monotonicity. Significant correlations here suggest that some monotonic relationship exists between two variables, which may not necessarily be linear. Significant correlations in this table which are not significant in Table~\ref{tab:correlation_spearman} may be modelled by non-linear assumptions. In this case, researchers seeking to model realistic angry driving may enforce constraints on learned trajectory curvatures based on correlation coefficients in driving policy objectives.}
\centering
\begin{tabular}{lc|ccc|ccccc}
\toprule
& & \multicolumn{3}{|c|}{Control Metrics} & \multicolumn{5}{|c}{Trajectory Metrics: Scenario 2}\\
Score & & $\sum$ Steer &  $\sum$ Throttle & $\sum$ Brake & Sinuosity & Curvature$_{TA}$ & Curvature$_{LV}$ & Curvature$_{Steiner}$ & Curvature$_{OC}$ \\
\hline\hline
\multirow{ 2}{*}{Reckless} & $r_{Spearman}$ &  0.338 & -0.154 & -0.456 & -0.078 & \textbf{0.620} & \textbf{0.603} & 0.429 & -0.240 \\
          & $p$ & 0.184 & 0.554 & 0.066 & 0.765 & \textbf{0.008} & \textbf{0.010} & 0.086 & 0.353 \\
\midrule 
\multirow{ 2}{*}{Anxious} & $r_{Spearman}$ & 0.199 & -0.164 & -0.444 & -0.002 & -0.35 & -0.314 & -0.164 & \textbf{-0.586} \\
          & $p$ & 0.445 & 0.529 & 0.074 & 0.993 & 0.168 & 0.22 & 0.529 & \textbf{0.013} \\
\midrule
\multirow{ 2}{*}{Risky} & $r_{Spearman}$ &  0.337 & -0.292 & -0.217 & -0.066 & 0.271 & 0.254 & 0.148 & 0.370 \\
          & $p$ & 0.186 & 0.256 & 0.403 & 0.801 & 0.293 & 0.326 & 0.57 & 0.143 \\
\midrule
\multirow{ 2}{*}{Angry} & $r_{Spearman}$ & -0.289 & 0.076 & 0.417 & 0.159 & \textbf{0.642} & \textbf{0.488} & \textbf{0.605} & 0.453 \\
          & $p$ & 0.260 & 0.772 & 0.096 & 0.541 & \textbf{0.005} & \textbf{0.047} & \textbf{0.010} & 0.068 \\
\midrule
\multirow{ 2}{*}{High-Velocity} & $r_{Spearman}$ & 0.025 & -0.098 & 0.265 & -0.189 & 0.35 & 0.265 & \textbf{0.522} & 0.108 \\
          & $p$ &  0.926 & 0.708 & 0.305 & 0.468 & 0.168 & 0.305 & \textbf{0.032} & 0.680 \\
\midrule
\multirow{ 2}{*}{Distress-Reducing} & $r_{Spearman}$ &  -0.017 & -0.355 & -0.115 & -0.27 & -0.419 & -0.35 & -0.402 & 0.010 \\
          & $p$ & 0.948 & 0.162 & 0.660 & 0.295 & 0.094 & 0.168 & 0.11 & 0.970 \\
\midrule
\multirow{ 2}{*}{Patient} & $r_{Spearman}$ & -0.473 & 0.353 & 0.444 & 0.088 & -0.135 & -0.184 & 0.108 & 0.017 \\
          & $p$ & 0.055 & 0.165 & 0.074 & 0.736 & 0.606 & 0.48 & 0.68 & 0.948 \\
\midrule
\multirow{ 2}{*}{Careful} & $r_{Spearman}$ & -0.022 & \textbf{0.564} & 0.130 & 0.174 & 0.039 & 0.105 & -0.154 & -0.110 \\
          & $p$ & 0.933 & \textbf{0.018} & 0.619 & 0.504 & 0.881 & 0.687 & 0.554 & 0.673 \\
\bottomrule
\end{tabular}
\label{tab:correlation_spearman}
\vspace{-1em}
\end{table*}

\subsection{Modeling Driving Personality with Driver Behavior}
\label{sec:method_model}

After data collection, we use the resulting questionnaire data and simulator sensor data to model driving personality with driving behavior. Our goal in this section is to be able to predict driver personality given the scenario encountered and the resulting trajectory of the vehicle. We model this relationship with a simple two-layer perceptron. This is for two reasons: 1) Our collected dataset is small, and thus no where near the size of datasets typically used for deep learning, and 2) the relationship between personality and driving behavior may be unintuitively complex, thus making simpler regression models difficult to fit. Two-layer perceptrons are a compromise between both ends of learning, and is supported by the Universal Approximation Theorem~\cite{Hornik_Stinchcombe_White_1989} in that our inputs are in a bounded and finite range (e.g., the bounded start and end route in each scenario). 

The inputs to the two layer perceptron is the spatio-temporal trajectory of vehicles in a particular scenario; four models are trained in total for each scenario. The trajectory $t \in \mathbb{R}^{T \times 2}$ is represented by a vehicle's position in two-dimensional Euclidean space, along $X$ and $Y$ axes, across $T$ timesteps. Each vehicle position in the trajectory is sampled at 20 Hz (20 frames per second). 

The output of our model is a vector of size 8, with each component representing the respective coefficient outputs of the Multidimensional Driver Style Inventory~\cite{Taubman-Ben-Ari_Mikulincer_Gillath_2004}: Reckless, Anxious, Risky, Angry, High-velocity, Distress-reducing, Patient, and Careful. The sum of each output vector is 1. 

The 2-layer perceptron has one hidden layer of 100 units, with ReLU function as the nonlinearity, dropout on the outputs of the first linear layer, and softmax applied on the output features to enforce sums to 1. We use early stopping to avoid overfitting during training, and stop training at a maximum of 200 epochs. Mean-squared error, or L2 error, is used as the loss function. 
We evaluate our model with k-fold cross validation, where $k=8$; evaluation results averaged across all folds are shown in Table~\ref{tab:kmeans_validation}.

\begin{figure*}[ht!]
    \centering
    \includegraphics[width=0.45\textwidth]{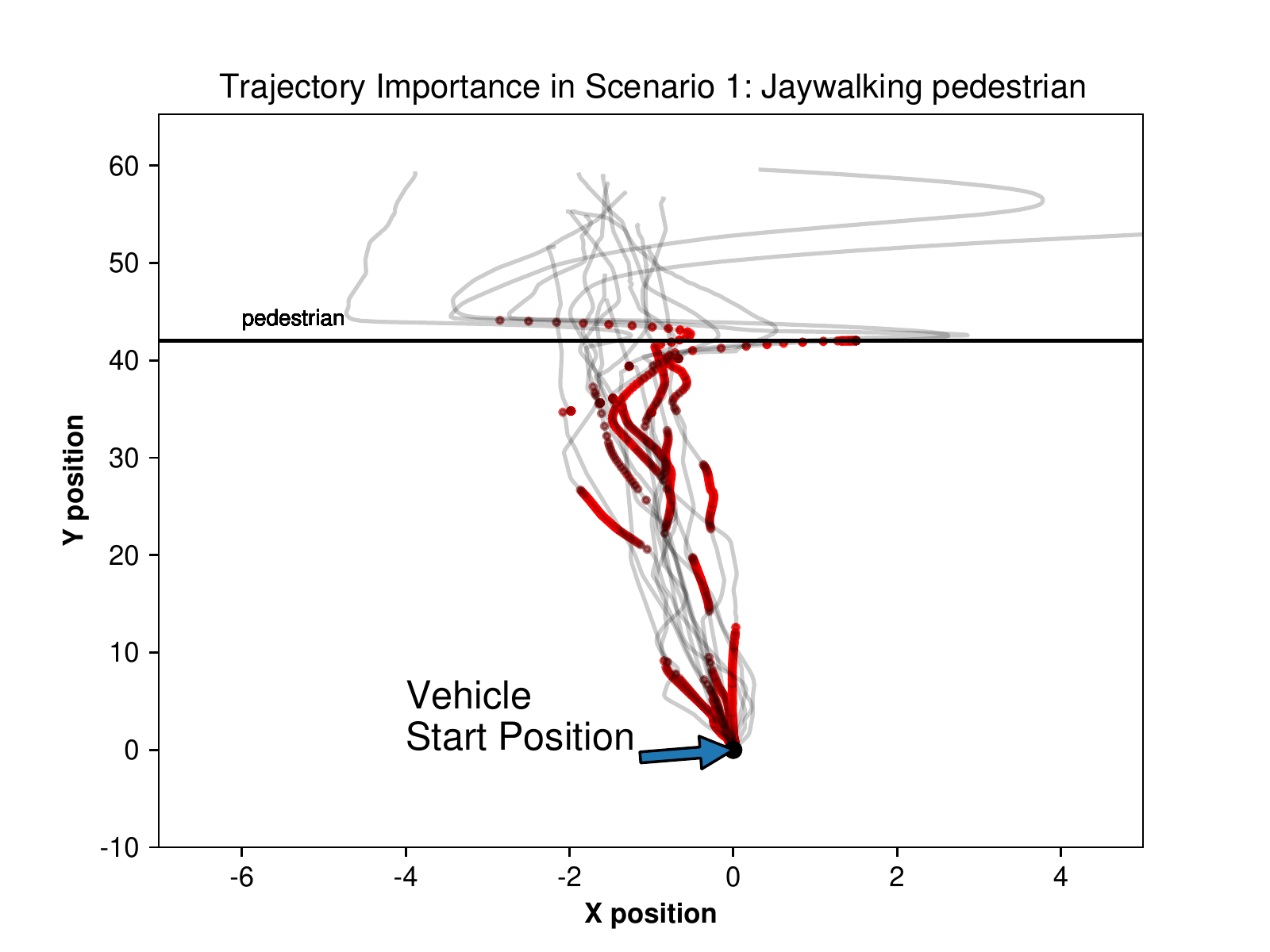}
    \includegraphics[width=0.45\textwidth]{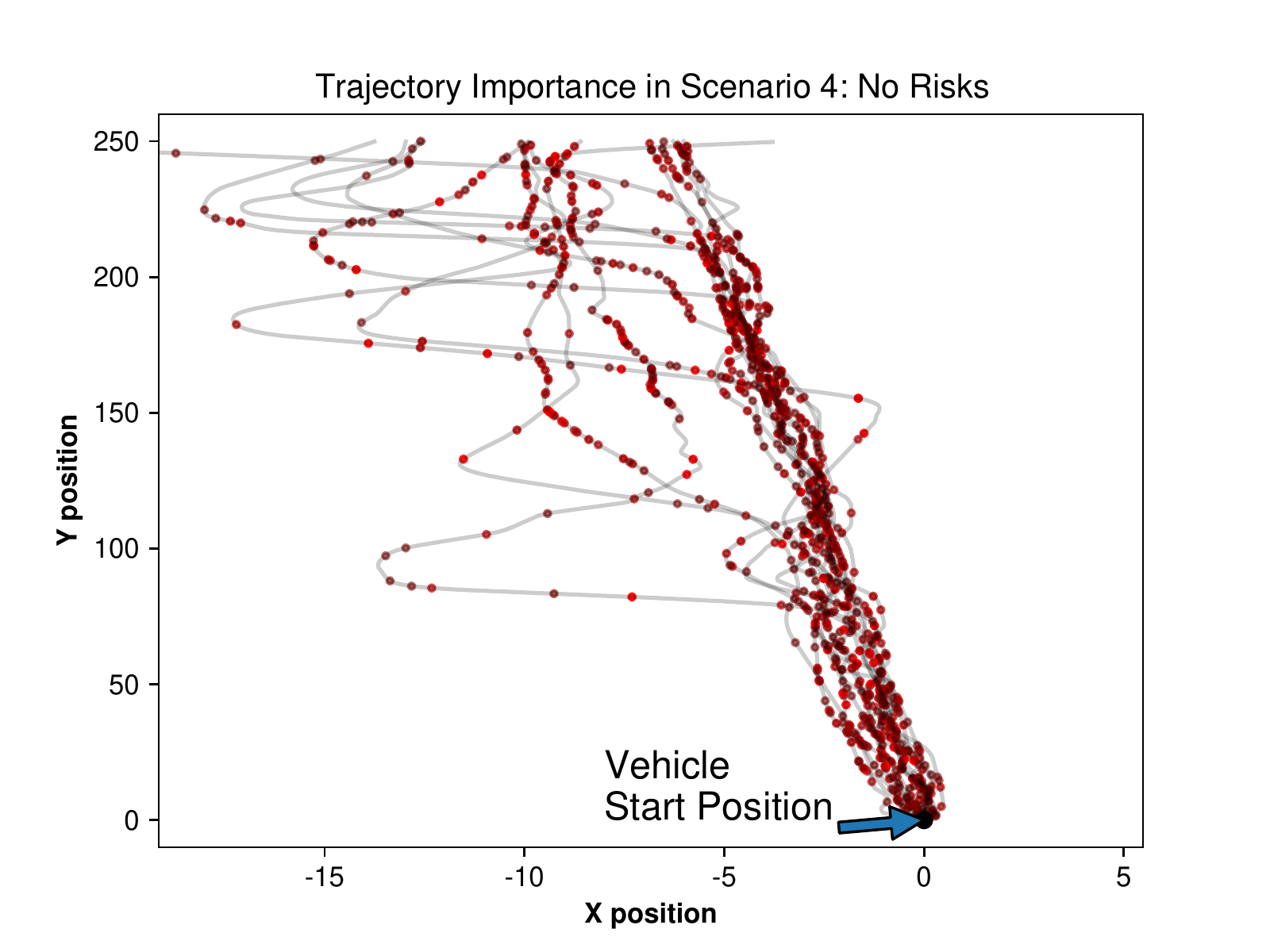}
    \caption{Trajectory importance for {\bf Scenario 1: Jaywalking Pedestrian (left)} versus {\bf Scenario 4: No risks (right)}. 
    The most important points of each user's trajectory in personality classification is highlighted in red. Salient regions for Scenario 1 are clustered at the beginning of the trajectory and around the area where the risk (pedestrian) is noticed. Conversely, in Scenario 4 with no risks, salient regions seem to be scattered randomly across entire trajectories. Neural networks are able to learn structured correlations along trajectories when risks are present. Intuitively, differences in personality may be more pronounced with a disruptive stimulus. This also shows that a trajectory history \textit{prior} to encountering the jaywalker may be enough to profile a driving personality, offering a prior for predicting future vehicle trajectory. 
    }
    \label{fig:heatmap}
    \vspace{-1.5em}
\end{figure*}

\section{Results}

In total, we collected data from 24 participants. The demographics of our sample population can be found in Table~\ref{tab:demographics}. We use the resulting dataset to analyze for relationships between driving behavior and personality, which can then be applied to driving policies for realistic personality simulation. 

\subsection{Correlation Analysis}

In this section, we use statistical methods to evaluate relationships in our collected data. Namely, we want to investigate the relationship between evaluated driving personality and driving behavior in our tested risky scenarios. For each personality component of MDSI, we compute its both Pearson and Spearman correlations \textit{to investigate whether a linear or monotonic relationship exists between personality factors and trajectory metrics.} 
Pearson and Spearman correlation results for Scenario 2 can be found in Tables~\ref{tab:correlation_pearson} and~\ref{tab:correlation_spearman} respectively. 
Pearson correlation tests for significant \textit{linear} relationships between variables, while Spearman correlation tests for significant monotonicity of relationships between variables. 
Variables with significant p-values for Pearson correlation tests will have a more rigorously defined linear relationship, while Spearman correlation will show whether a significant monotonic relationship exists between variables at all, even if it is not linear. We perform both tests for completeness.

We present Scenario 2 as it has the highest number of relationships where $p \leq 0.05$. In particular, we can observe an inverse correlation between patience and change in steering, a positive correlation between patience and change in braking, and a positive correlation between carefulness and change in throttle. While the patience correlations make intuitive sense, it is interesting to note that increasing carefulness is correlated with increasing throttle. This may be due to careful drivers pressing the throttle lightly over a prolonged number of timesteps to accelerate, rather than pressing hard for a brief moment.


While most variable pairs do not have a p-value below the 0.05 threshold, the correlation coefficients and p-values are still directly applicable to driving policies. Correlation coefficients describe the relationship between variables, while p-values describe the confidence of the relationship model. Thus, a specific driving personality can be modeled as a linear of function of attainable driving data. For example, a control policy seeking to imitate \textit{patient} driving can enforce linear constraints on its actions based on steering, braking, and throttle. P-value can then be used to encode importance of each term in the constraint. 

Significant correlations between personality scores and trajectory metrics, located in the right half of both tables, are interesting but less intuitive. Different measures of trajectory are borrowed from fundamental differential geometry literature. In particular, we compute sinuosity, Turning Angle Curvature (Curvature$_{TA}$), Length Variation Curvature (Curvature$_{LV}$), Steiner Curvature (Curvature$_{Steiner}$, and Osculating Circle Curvature (Curvature$_{OC}$). Curvature metrics quantify the spatial ``curviness" of a discrete trajectory; we use five curvature definitions since each one captures different characteristics of point-based, non-differentiable trajectories. Across all curvature metrics, we observe correlations between trajectory and anxiousness, riskiness, angriness, high-velocity driving, and distress-reducing behavior. The presence of significant correlations in specific scenarios are indicative of some sort of correlations between path taken and personality factors. 

One limitation of these trajectory metrics is that the time dimension is collapsed, and only the shape of the resulting trajectory is considered. It could be that two different users followed similar paths, but in a different amount of time. 
Control input metrics do however account for the time dimension since values are summed over all timesteps; longer trajectories will generally indicate higher values for control input sums. 
We account for this with the control input results, which are directly related to vehicle movement over time. 

    

\subsection{Modeling Perceived Personality of Driving Behavior}

Alternatively, we can model the relationship between driving data and personality with the neural network described in Section~\ref{sec:method_model}, and analyze how networks learn from trajectory data. Since the point of collecting risky data is to better identify and avoid risks, we want to know if personality information can help neural networks encode trajectory information in risk encounters. 
We visualize the most important points on a trajectory in personality classification for scenarios 1 and 4 in Figure~\ref{fig:heatmap}. In other words, we want to answer ``What parts of the driving trajectory in risky situations are most indicative of a driver's personality?". Red colors indicate trajectory positions in which the gradient of the output with respect to the inputs was in the 90$^{th}$ percentile of all sampled positions. 
In Scenario 1, an initially occluded pedestrian jaywalks in front of the user once the user drives into a fixed trigger location. We can observe that, on the left in Figure~\ref{fig:heatmap}, the model considered the  the beginning of the trajectory and the preceding 20 meters before the jaywalker crossing as most indicative of personality classification. In contrast, the control scenario on the right side of Figure~\ref{fig:heatmap} shows no particular patterns in the 90$^{th}$ percentile of gradients, with no obvious qualitative correlations between trajectory position and personality classification. This result qualitatively shows that personality information is densely encoded into small structured segments in the jaywalking scenario, whilst personality information is encoded randomly across trajectories in a scenario with no risks; this suggests that neural networks are capable of learning intuitive and structural characteristics of driving behavior when risks are present in the scenario. 
The pattern of trajectory importance in the jaywalking pedestrian scenario is interesting, since personality indicators are most densely encoded in positions leading up to the jaywalker's crossing path. This suggests that a driver's trajectory prior to encountering a jaywalking pedestrian is descriptive enough to indicate how they may drive in the future, via personality profiling.



\subsection{Evaluation on Immersion and Simulator Sickness}
While our results demonstrate some distinct behaviors for different personalities of drivers, it is important to note that the user behavior can be affected by features unique to immersive VR simulators.
First, since participants are not actually driving a real car, their safety is guaranteed even in the risky scenarios we collected data on. Thus, it is possible that their behaviors in the simulation will not closely mimic their real-world behaviors. However, we believe this is not a large factor in our work because 79\% of participants reported high levels of immersion during the driving tasks.
Second, some participants may experience simulator sickness (a type of motion sickness that is unique to immersive simulations like ours), which can potentially influence their behaviors.
In our data,  only 7\% of participants experienced severe simulated sickness. 
Additional participant sentiments, demographics, and visualization of simulator sickness symptoms before and after the VR portion of the user study can be found on the project website. 



\subsection{Qualitative Observations}

Various participants asked if they could adjust the front mirror; this illustrates that some participants have routinized behaviors before they start driving a car. A few participants even instinctively grasped for a nonexistent physical gear stick or lane change controls, which suggests high levels of immersion in the simulator.
Additionally, although some participants of our study used the car mirrors in the simulator to check whether there were cars in their surroundings before changing lanes, most of them did not use them. This could be due to the fact that they knew they were in a simulator so their lives were not at-risk with those maneuvers.
Moreover, one participant highlighted that they wished they could have seen both their legs and the pedals in the simulator.
Finally, in one scenario there were bumps on the street; only one participant noticed them.


\subsection{Limitations and Discussion}
Our study has some limitations. 
For instance, the simulator setup did not implement a car horn or turn signals, which are elements that drivers commonly use in the real world. In addition, the driving simulator situated the driver on the left side of the vehicle, and all participants were used to that setup when they drive real cars. We acknowledge that in some countries the driver's seat is on the right-side; none of our participants are used to that scheme based on their driver's license identification. 
The driving simulator, though ported to VR, is typically used as an autonomous driving testbed, and thus is not developed with VR user experience in mind. Due to this, frame rate may degrade in some scenarios with denser traffic and lead to higher degrees of simulator sickness symptoms. With more advancements in driving simulation with VR user studies in mind, the efficiency and experience of data collection can be greatly improved, along with opportunity with diverse scenarios. 

Interesting angles of future work can also be explored following the results from Figure~\ref{fig:heatmap}. Driver personality may be densely encoded in moments leading up to a disruption from typical, predictable driving behavior. The neural network consistently assigns high gradient values to positions preceding the jaywalking path, indicating some sort of variance. Since salient segments are also short, it may be possible to efficiently and quickly predict the trajectory of a nearby vehicle. This can help autonomous driving policies predict safe collision avoidance in complex chain reactions in the wild. Furthermore, simulated non-human drivers can also represent realistic personality behaviors by using correlations from Tables~\ref{tab:correlation_pearson} and~\ref{tab:correlation_spearman} as constraints on control inputs and planned waypoint trajectories. The same can also be applied for autonomous driving policies, eliminating the need for collecting a large dataset from human subjects. 
\section{Conclusion}
In this paper, we conducted the first VR driving user study to identify how driver personalities relate to measured behaviors in risky situations for autonomous driving, and validated our collected data with analysis and variable relationship modeling. Our work has several implications: 1) our work can serve as a reference for the autonomous driving community interested in conducting human subject data collection or user studies, as VR user studies in autonomous system research are sparse and, to the best of our knowledge, unprecedented, 2) incorporating quantifiable personality characterizations into autonomous driving research can lead to interpretable driving for both the researchers and the end users, and 3) insights from human subject simulation data in risk scenarios can be applied to both autonomous and non-human driving behavior without the need for large datasets or deep learning.  Such a modeling framework can be further extended to incorporate various uncertainties in sensor data and/or changes in environmental factors (e.g. lighting and weather conditions).

\section{Acknowledgements}
We would like to thank Ian McDermott, Roger Eastman, and the Immersive Media Design Lab at the University of Maryland for their generosity in lending their time, lab space, computers, and VR hardware. Additionally, we thank Niall Williams for advice on VR study design, analysis, and lending us a pupillometer.

\bibliographystyle{IEEEtran}
\bibliography{ref}

\end{document}